%% file: paper_main.tex
\pdfoutput=1

\documentclass[11pt]{article}

\usepackage[final]{acl}

\usepackage{times}
\usepackage{latexsym}

\usepackage[T1]{fontenc}

\usepackage[utf8]{inputenc}

\usepackage{microtype}

\usepackage{inconsolata}

\usepackage{graphicx}
\usepackage{booktabs}
\usepackage{amssymb}
\usepackage{amsmath}
\usepackage{comment} 
\usepackage{float}
\usepackage{makecell}
\usepackage{cleveref}

\title{Controlled Generation for Private Synthetic Text}

\author{Zihao Zhao \\
  Johns Hopkins University \\
  \texttt{zzhao71@jhu.edu} \\\And
  Anjalie Field \\
  Johns Hopkins University \\
  \texttt{anjalief@jhu.edu} \\}

\begin{document}
\maketitle
\begin{abstract}
Text anonymization is essential for responsibly developing and deploying AI in high-stakes domains such as healthcare, social services, and law.
In this work, we propose a novel methodology for privacy-preserving synthetic text generation that 
leverages the principles of de-identification and the Hiding In Plain Sight (HIPS) theory. Our approach introduces entity-aware control codes to guide controllable generation using either in-context learning (ICL) or prefix tuning. The ICL variant ensures privacy levels consistent with the underlying de-identification system, while the prefix tuning variant incorporates a custom masking strategy and loss function to support scalable, high-quality generation. Experiments on legal and clinical datasets demonstrate that our method achieves a strong balance between privacy protection and utility, offering a practical and effective solution for synthetic text generation in sensitive domains.\footnote{The code is provided in https://github.com/zzhao71/Controlled-Generation-for-Private-Synthetic-Text.git}

\end{abstract}

\section{Introduction}
Text anonymization is an essential precursor to responsibly developing, deploying, and auditing AI in high stakes domains, like healthcare \citep{panchbhai2022systematic}, social services \citep{gandhi2023annotating}, or law \citep{zhong2020does}, where sharing sensitive data with AI developers or using it to train models risks harmful leakage.
Despite its importance, achieving effective anonymization in text is notoriously challenging. Tools for redacting directly identifying content, like names and addresses, have grown increasingly accurate, but they are unlikely to guarantee 100\% recall, leaving ``residual'' identifiers easy to spot \citep{carrell2013hiding}.
They also fail to remove vaguer quasi-identifying information that can still lead to re-identification, like a description of someone's appearance \citep{lison2021anonymisation,Pilán2022}.

Recently, synthetic text has become a potentially more robust alternative to long-standing redaction approaches. The idea involves leveraging Large Language Models' (LLMs) ability to generate highly fluent outputs to create text that is realistic enough to be useful but scrambles and abstracts away identifying or sensitive information. Models are typically fine-tuned over the real data and then prompted to generate new text. As LLMs are prone to memorizing and outputting training data \citep{carlini2021extracting}, prior work has conducted fine-tuning using differential privacy to prevent leakage of sensitive information \citep{yue2023synthetic, kurakin2023harnessing, mattern2022differentially, putta2023differentially}. However, although language-model-generated data quality is generally high in the above methods, differentially private fine-tuning greatly degrades synthetic text quality \citep{ramesh-etal-2024-evaluating}. Furthermore, differential privacy guarantees are difficult to maintain in text settings where the unit of privacy is often unclear, and in practice, they do result in leakage of directly identifying information \citep{ramesh-etal-2024-evaluating}.

In this work, we propose a methodology for privacy-preserving synthetic data generation that aims to balance data utility with protection against the leakage of sensitive identifiers. Rather than relying on differential privacy, our approach draws on the long-established practice of de-identification and the theory of Hiding In Plain Sight (HIPS), which suggests that replacing detected identifiers with realistic surrogates can obscure the presence of any leaked real identifiers, making them more difficult to detect or exploit \citep{hirschman2010measuring, carrell2013hiding}.

Our method begins by identifying private entities within the input text and representing them as control codes. These control codes are then used to guide controllable text generation \citep{keskar2019ctrl}, where falsified sensitive information is injected into the generation process to reduce the likelihood of the model reproducing real identifiers. We propose two variants of this approach: an in-context learning (ICL) method and a prefix tuning-based fine-tuning method.

In the ICL setting, the model is prompted with several example contexts, each consisting of a control code and its corresponding private text, followed by a fictional control code to guide the generation of the synthesized text. We further enhance privacy by directly blocking the output of sensitive tokens from the private examples, thus providing at least the same level of privacy protection as the underlying de-identification system (expecting small easy-to-filter errors due to case sensitivity). This characteristic makes the system suitable for deployment in regulated domains where legal compliance, such as with HIPAA, is essential.

\begin{figure*}[t]
  \centering
  \includegraphics[width=1.0\linewidth, height = 0.2\linewidth]{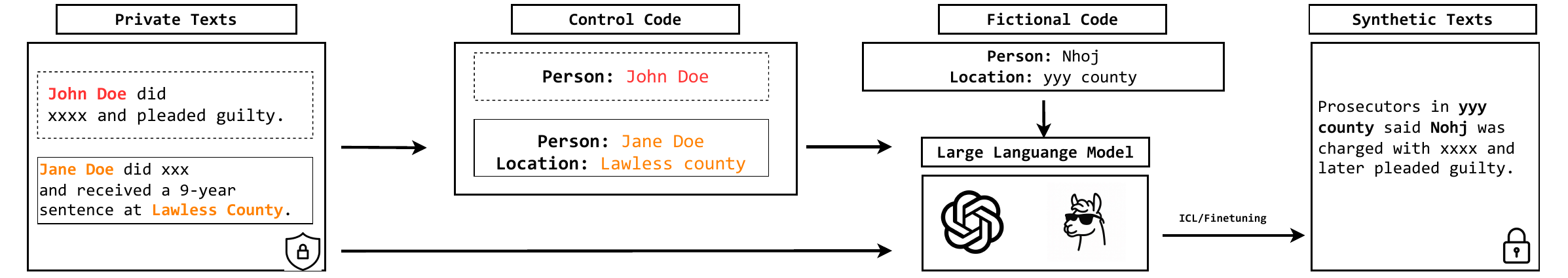}
  \caption{Overview of synthetic data generation}
  \label{fig:example_figure}
\end{figure*}

In the prefix tuning variant, the model is fine-tuned using input-output pairs composed of control codes and their associated private texts. A fictional control code is again provided for the generation of synthetic text. This approach is further enhanced by our proposed privacy masking and custom loss function, which improve privacy protection and enable the generation of high-quality synthetic data. The overall method is illustrated in Figure~\ref{fig:example_figure}.

We conduct experiments over data from two domains with sensitive information: law \citep{Pilán2022} and healthcare \citep{johnson2016mimic}, demonstrating that our approach successfully produces synthesized text, prevents leakage of identifiers, and maintains text utility for downstream tasks. Unlike current paradigms that rely on DP-SGD for model fine-tuning \citep{yue2023synthetic}, our novel method achieves a better balance between privacy protection and utility while being significantly more efficient and practical for real-world applications. This approach offers strong potential for advancing the responsible development and deployment of AI in contexts where both data utility and privacy preservation are critical considerations.


\input{paper/method}
\input{paper/experiment}
\section{Related Work}

\paragraph{Synthetic data}
Synthetic data refers to the data that is generated artificially instead of collecting real-world events or annotations. In the context of AI, it specifically refers to the data that is generated by generative models to mimic the characteristics of real data \citep{liu2024best, saxton2019analysing}.

In NLP, LLMs have replaced more traditional methods for synthetic data generation, like synonym replacement, random insertion, and back-translation, and are emerging as a potentially viable alternative to human-generated data \citep{hartvigsen2022toxigen, ye2022zerogen}. 
Extensive pretraining allows LLMs to generated seeming fluent outputs \citep{ding2023gpt}, with a high level of controllability and adaptability, allowing researchers to create flexible datasets tailored to specific requirements \citep{eldan2023tinystories}.

To ensure privacy in synthetic data generation, differential privacy (DP) has emerged as a foundational framework \citep{dwork2006calibrating}. Early efforts applied DP to generative models such as generative adversarial networks (GANs) and variational autoencoders (VAEs) by modifying training procedures using differentially private stochastic gradient descent (DP-SGD) \citep{xie2018differentially, chen2020gs, abadi2016deep}. More recently, DP has been extended to language models and text generation. For instance, \citet{lilarge}, \citet{yu2021differentially}, and \citet{ramesh-etal-2024-evaluating} fine-tune large language models under DP constraints by incorporating gradient clipping and noise injection to mitigate the memorization of sensitive content.

Our approach is complementary to fully differentially private training. Rather than enforcing global privacy guarantees, we focus on entity-aware generation through in-context learning and prefix tuning. This strategy reduces the risk of private information leakage by guiding the model to operate on synthetic or fictional identifiers, without the utility degradation imposed by DP. 

\paragraph{Text Sanitization}
Text sanitization refers to the process of replacing sensitive tokens to protect privacy \citep{tong2025vulnerability}. A common approach involves first identifying text spans that contain personally identifiable information (PII) and then replacing them with a default placeholder, such as '***' \citep{olstad2023generation} or a black box \citep{lison2021anonymisation, pilan2022text}. However, this method often reduces the utility of the text. To mitigate this issue, alternative strategies have been developed that replace sensitive content with less risky alternatives, such as synonyms \citep{dalianis2019pseudonymisation, volodina2020towards} or synthetic surrogates \citep{carrell2013hiding}. In the medical domain, for example, patient names may be substituted with randomly selected names from a predefined list \citep{dalianis2019pseudonymisation}.

In addition, differential privacy (DP) has been applied to text sanitization at both the word level and sentence level. Word-level DP methods perturb individual words to achieve privacy guarantees \citep{feyisetan2020privacy, feyisetan2019leveraging}, while sentence-level approaches ensure privacy across entire documents \citep{igamberdiev2023dp, krishna2021adept, mattern2022limits, utpala2023locally}. These DP-based techniques offer formal privacy guarantees for each record in a dataset. However, such methods often come at the cost of degraded text quality and reduced utility. In contrast, our approach strikes a more favorable balance between privacy protection and text utility, generating high-quality synthetic text while mitigating privacy risks.

\section{Conclusion}
In our work, we proposed two approaches for generating privacy-preserving synthetic text that aims to balance privacy protection with downstream utility: an ICL method and a prefix tuning-based fine-tuning method. While the ICL method offers stronger privacy protection—particularly when combined with the privacy enhancement mechanism—it still exhibits notable lexical overlap with the original private text and has lower utility. In contrast, the prefix tuning with masking method achieves a better trade-off, demonstrating strong performance in both privacy and utility. Although it shows slightly higher entity-level leakage, it produces more diverse and less memorized outputs, which have high applicability in real-world scenarios.

Overall, both approaches demonstrate strong potential for use in synthetic text generation under privacy constraints. In future work, we plan to evaluate these methods on a broader range of datasets and explore additional hyperparameter settings to further improve performance.

\section{Acknowledgments}
We thank reviewers for their helpful feedback. This work was supported in part by the AI2050 Fellowship program by Schmidt Sciences.

\section{Limitations}
Our current study has several limitations. First, we evaluate our methods on only two datasets, which limits the generalizability of our findings. Future work should include a broader range of datasets across different domains and scales to assess performance more comprehensively. Second, we use fixed hyperparameters in our experiments. A more extensive hyperparameter search—including variations in the number of in-context examples and the loss weighting coefficients $\lambda_{\text{LM}}$, $\lambda_{\text{contrastive}}$, and $\lambda_{\text{KL}}$—could further improve performance. We also fix decoding parameters such as top-$p$ and temperature, which may not be optimal across all settings. Finally, all experiments are conducted using the Sheared LLaMA 1.3B model. Evaluating our methods on larger range of language models would help determine how performance varies and whether models exhibit different privacy-utility trade-offs.
\bibliography{custom}
\input{paper/appendix}

\end{document}

%% file: paper/method.tex
\section{Methodology}
Given a corpus of documents $D_{\text{real}}$ containing sensitive information—such as court cases or clinical notes—our objective is to construct a synthetic dataset $D_{\text{synthetic}}$ that preserves the core content and stylistic characteristics of $D_{\text{real}}$ while preventing the leakage of private information.

To achieve this, we employ \textit{control codes} to guide the generation process toward producing fabricated sensitive information rather than reproducing real identifiers.
In general, control codes have been used to guide models to produce outputs with desired properties  \citep{keskar2019ctrl}. Under this framework, both training and inference rely on the conditional probability of the output sequence $x = {x_1, x_2, \ldots, x_n}$ given the control code $c$, defined as:

\begin{equation}
P(x \mid c) = \prod_{i=1}^{n} P(x_i \mid x_{<i}, c)
\label{eq:ctrl}
\end{equation}

We first define our notion of control codes and then how we use them to facilitate generation through either in-context learning or fine-tuning.

\subsection{Control Codes}

We define control codes as categories of private information (e.g., names, dates, or locations) and their associated values.

For a real document $d \in D_{real}$, which serves as an ICL example or a training data point, we extract all associated identifiers, such as names, ID numbers, or addresses—using either  manual annotation or a de-identification model \citep{MsPresidio}.
We group all identifiers by type to construct a control code of \texttt{<IDENTIFIER TYPE>: <REAL\_VALUE>, <REAL\_VALUE>,...} pairs that contains all sensitive information in $d$. For example, $c=$

\texttt{PERSON: Alice Jones, John Smith}

\texttt{LOC: New York City}

\texttt{...}

Accordingly, every sensitive span in $d$ is associated with one or more identifier types.
The primary role of control codes is to help the model identify sensitive content and learn its relationship to the surrounding context.

At inference time, we construct fictional control codes—synthetic representations that follow the same format but contain falsified values randomly sampled from public lists. These fictional control codes direct the model to generate fake sensitive information over real information in accordance with the HIPS theory \citep{hirschman2010measuring, carrell2013hiding}.


\subsection{In-context learning (ICL)}
In the ICL approach, we sample three distinct real documents $d_{1:3}$ from $D_{\text{real}}$. We construct associated control codes $c_{1:3}$, and a fictional code $c_f$. We then prompt an LLM with the three real pairs followed by the fictional code: $<c_1,d_1>$, $<c_2,d_2>$, $<c_3,d_3>$, $c_f$. Thus, the LLM is prompted to generate a synthetic document consistent with the preceding examples but grounded only in the fictional control code, thereby avoiding the use of any real sensitive data. Examples of contexts in in-context learning generation are shown in Appendix~\ref{in-context-learning}.

\paragraph{ICL w/ privacy enhancement}
In the ICL setup, the set of private information that the generation model is exposed to and could possible reproduce is limited to the small finite set of values in $c_{1:3}$.
We explicitly designate these values as a set of ``bad'' tokens that the language model is prohibited from generating.\footnote{We use \texttt{AutoModelForCausalLM.generate(..., bad\_words\_ids=...)} to hard-block specific tokens at decoding time.} During decoding, these tokens are assigned low or zero probability, effectively preventing the model from selecting them in the output. This approach enforces that sensitive content is excluded from the model's output, ensuring that privacy preservation is at least as good as the hand annotations or deidentification model used to construct  $c_{1:3}$.

\subsection{Fine-tuning}
In the fine-tuning setup, we train the LLM to generate a document from a provided control code. More specifically, from $D_{real}$, we construct training pairs where $c_i$ is the input and $d_i$ is the output and fine-tune the LLM over these pairs. We employ prefix-tuning \citep{li2021prefix}, which is much more computationally efficient than full fine-tuning and has been shown to successfully adapt models to similar tasks \citep{he2023large}. At inference time, fictional codes $c_f$ are used as inputs to direct the model to output documents with fictional sensitive information.

\paragraph{Fine-tuning w/ Masking}
Inspired by \citet{he2023large}, we introduce a masked loss framework to enhance privacy protection during prefix tuning. We define a binary mask $M$ to indicate the privacy status of each token in the training text. Specifically, private tokens are masked with $M = 0$, while non-private tokens are masked with $M = 1$.

Let $P_\theta$ denote the fine-tuned model’s probability distribution. The standard language modeling loss $\mathcal{L}_{\text{LM}}$ is defined as:

\begin{equation}
\mathcal{L}_{\text{LM}} = -\sum_t \log P_\theta(y_t \mid y_{<t})
\end{equation}

where $y_t$ is the ground truth token at position $t$, and $y_{<t}$ denotes all preceding tokens before position $t$.

To further enforce a divergence in behavior on private spans, we define a contrastive loss $\mathcal{L}_{\text{contrastive}}$, which encourages the fine-tuned model $P_\theta$ to deviate from the base model $P_{\text{base}}$ specifically on private tokens:

\begin{equation}
\scalebox{1.1}{$
-\sum_t (1 - M_t) \log \left(
    \frac{P_\theta(y_t \mid y_{<t})}
    {P_\theta(y_t \mid y_{<t}) + P_{\text{base}}(y_t \mid y_{<t})}
    \right)
$}
\end{equation}

This loss focuses only on private tokens ($M_t = 0$), promoting prediction divergence between $P_\theta$ and $P_{\text{base}}$.

In contrast, to preserve utility on non-private content, we introduce a KL divergence loss $\mathcal{L}_{\text{KL}}$, which encourages the fine-tuned model to remain close to the base model on non-private tokens:

\begin{equation}
\scalebox{1.0}{$
\mathcal{L}_{\text{KL}} = \sum_t M_t \sum_{v \in V} P_{\text{base}}(v \mid y_{<t}) \log \frac{P_{\text{base}}(v \mid y_{<t})}{P_\theta(v \mid y_{<t})}
$}
\end{equation}

where $V$ denotes the vocabulary and $v$ represents a token in $V$.

Finally, the overall training objective is a weighted combination of the three losses:

\begin{equation}
\mathcal{L}_{\mathrm{total}} = 
\lambda_{\mathrm{LM}} \mathcal{L}_{\mathrm{LM}} + 
\lambda_{\mathrm{contrastive}} \mathcal{L}_{\mathrm{contrastive}} + 
\lambda_{\mathrm{KL}} \mathcal{L}_{\mathrm{KL}}
\end{equation}

The hyperparameters $\lambda_{\text{LM}}$, $\lambda_{\text{contrastive}}$, and $\lambda_{\text{KL}}$ control the relative contributions of each component.

%% file: paper/experiment.tex
\section{Experimental setup} 
\subsection{Dataset} We conduct our experiments using the Text Anonymization Benchmark (TAB)  introduced by \citet{Pilán2022}, which comprises English-language court cases from the European Court of Human Rights (ECHR) and is manually annotated with personal identifiers. We use both the training (1,014 entries) and test (127 entries) splits. The training set is primarily used for synthetic data generation, while the test set is employed to evaluate the utility of models fine-tuned on the synthetic data.

Following the annotation scheme proposed by \citet{Pilán2022}, we categorize information into three types: direct identifiers, quasi-identifiers, and non-sensitive information. Direct identifiers refer to information uniquely associated with an individual, whereas quasi-identifiers are publicly known attributes that may not independently lead to re-identification but could do so when combined with other such attributes. In this work, we define sensitive information as direct identifiers. Due to the model's input token limitations, we split each document into two segments: the first 6 paragraphs and paragraphs 6 to 12. The reported results represent the average scores across these two segments.  

For experiments in the healthcare domain, we utilize the widely used dataset: MIMIC-III \citep{johnson2016mimic, goldberger2000physiobank}. The MIMIC-III dataset contains over 2 million de-identified clinical notes associated with more than 40,000 patients admitted to the Beth Israel Deaconess Medical Center in Boston, Massachusetts. To generate synthetic data from MIMIC-III, we use the string used for deidentification (e.g., [**Hospital 18**]) as control codes. The dataset provides diverse  medical text for evaluating the effectiveness of our privacy-preserving synthetic generation methods in high-stakes domains.

\subsection{Model}
For both synthetic text generation and fine-tuning with synthetic data to evaluate utility metrics, we use the Sheared-LLaMA 1.3B model \citep{xiasheared}.
In the in-context learning setting,
The generation parameters are set with a maximum of 400 new tokens, a temperature of 0.7, and a top-$p$ value of 0.9.
For prefix tuning, we use a default of 20 virtual tokens in the prefix. The model is trained for 3 epochs with a learning rate of $5 \times 10^{-5}$. The loss weighting coefficients are set as follows: $\lambda_{\text{LM}} = 1$, $\lambda_{\text{contrastive}} = 1$, and $\lambda_{\text{KL}} = 1$, assuming equal importance across all components.\footnote{We did not perform hyperparameter tuning; it is possible that further tuning could improve the method's performance.}

\subsection{Control Code Construction}

We evaluate models under two settings for constructing control codes. In the first, \textbf{private entity known}, we use the 
expert annotations of private entities provided in TAB, which are categorized into several types, including \textit{person}, \textit{code}, \textit{location}, \textit{organization}, \textit{demographic}, \textit{datetime}, \textit{quantity}, and \textit{miscellaneous}. A detailed description of these categories is provided in Appendix~\ref{appendix_tab_control_code}. For the MIMIC dataset, which has already been anonymized, the anonymized terms will be recognized as control codes.

In the second, \textbf{private entity unknown}, we consider how our method might perform for anonymizing a dataset that has not already been annotated by experts for sensitive information, which is a more realistic scenario. In this case, we employ an existing entity recognition tool,  Presidio \citep{MsPresidio}, to automatically identify potential private entities.
We construct control codes using these inferred sensitive spans rather than the hand-annotated ones.
In both settings, we use the hand-annotated sensitive spans in evaluation privacy leakage.


We also construct fictional control codes the same way in both settings.
Fictional control codes are randomly generated, and the full details of the generation procedure are provided in Appendix~\ref{fictional_control_code}.

\subsection{Privacy Metrics} 
We introduce several metrics to quantify privacy leakage. 

\paragraph{Private Information Presence Percentage (PIPP)}
We define \textit{Private Information Presence Percentage (PIPP)} as the proportion of generated passages that contain at least one private entity. In the in-context learning setting, a privacy leak is recorded if any private term from the provided context appears in the generated output. In the prefix-tuning setting, a generation is considered a leak if any private entity from the entire fine-tuning dataset appears in the output. This metric quantifies the frequency of such leaks relative to the total number of generated samples.

Let $s_1, \dots, s_n$ denote the $n$ generated outputs. The PIPP is computed as:

\begin{equation}
\text{PIPP} = \frac{1}{n} \sum_{i=1}^{n} \mathbb{I}(s_i \text{ contains private information}) 
\end{equation}

where $\mathbb{I}(\cdot)$ is the indicator function, returning 1 if the condition is true and 0 otherwise.

\paragraph{Entity Leakage Percentage (ELP)}
The \textit{Entity Leakage Percentage (ELP)} measures the proportion of private entities that appear in the model’s generated output. In the ICL setting, we count how many private entities from the input context are present in the synthetic output, then divide this count by the total number of private entities in the context. We report the average value of this ratio across all examples.

In the fine-tuning setting, we first identify the total number of deduplicated private entities in the training set, as the model is exposed to all of these entities and could potentially leak them. We then collect a deduplicated list of leaked private entities found in the generated text. The ELP is computed as the ratio of the number of leaked (non-repetitive) entities to the total number of private entities.


\paragraph{Rouge Score}
While PIPP and ELP both measure leakage of identifiers, the goal of generating synthetic text instead of just redacting identifiers is for the synthetic text actually to be synthetic: e.g., it should contain new combinations of words and information that mask any residual identifiers, rather than directly copying content from model inputs.
In order to evaluate if the generated text is sufficiently synthetic, we also report ROUGE-2 and ROUGE-L scores \citep{lin2004rouge} to evaluate the similarity between the generated text and the reference text. The specific formulas for both ROUGE-2 and ROUGE-L scores are presented in Appendix~\ref{rouge_score}.

In the in-context learning setting, the reference text corresponds to the context provided to the model during generation. For the fine-tuning method, we generate each fictional control code for each fine-tuning sample sharing the same identifier type. To check if the output is copying from its fine-tuning data, the ROUGE score is then computed between the model's output and the corresponding text in the fine-tuning dataset, and takes the maximum value.
Higher ROUGE scores in this context indicate a greater resemblance between the generated text and the original private content, and thus a higher risk of privacy leakage.
\begin{table*}[t!]
  \centering
  \begin{tabular}{l|cccc}
    \toprule
    \textbf{Method} & \textbf{PIPP (\%)}$\downarrow$ & \textbf{ELP (\%)} $\downarrow$& \textbf{ROUGE-2} $\downarrow$&\textbf{ROUGE-L} $\downarrow$ \\
    \midrule
    Baseline & 50.0 $\pm$ 2.20 & 30.89 $\pm$ 3.15 & 0.3898 $\pm$ 0.02 & 0.4303 $\pm$ 0.01 \\
    DP-SGD & 1.56 $\pm$ 0.02 & 0.73 $\pm$ 0.05 & 0.0635 $\pm$ 0.002 & 0.1328 $\pm$ 0.01	\\
    ICL & 3.25 $\pm$ 0.23 & 0.36 $\pm$ 0.012  & 0.2978 $\pm$ 0.02 & 0.3073 $\pm$ 0.008\\
    ICL w/ privacy enhancement & \textbf{0.00 $\pm$ 0.00} & \textbf{0.00 $\pm$ 0.00} & 0.3361 $\pm$ 0.04 & 0.3645 $\pm$ 0.005 \\
    Fine-tuning  & 3.94 $\pm$ 0.2 & 0.35 $\pm$ 0.07 & 0.017 $\pm$ 0.001& 0.133 $\pm$ 0.004\\
    Fine-tuning w/ masking & 2.56 $\pm$ 0.89 & 0.39 $\pm$ 0.03 & \textbf{0.0111 $\pm$ 0.003} & \textbf{0.0975 $\pm$ 0.002}\\
    \bottomrule
  \end{tabular}
  \caption{Privacy protection performance under the private entity known setting on the TAB dataset}
  \label{tab:private_entity_known}
\end{table*}
\subsection{Utility Metrics}
In addition to privacy protection, utility is also an important metric for measuring whether the content of sentences or paragraphs is preserved.
\paragraph{Perplexity}
After generating the synthetic text, we fine-tune a new instance of the Sheared LLaMA 1.3B model using the synthetic data as training input. To evaluate the utility of the synthetic data, we measure the model's perplexity on the test set \citep{jelinek1998statistical}. Perplexity is a commonly used metric in language modeling that quantifies how well a model predicts a sequence of tokens. It is computed as the exponential of the average negative log-likelihood of the true tokens under the model's predicted distribution. Lower perplexity indicates better predictive performance, suggesting that the synthetic data supports effective learning and retains high utility for downstream language modeling tasks.

\paragraph{MAUVE}
MAUVE (short for Model and Human Outputs Via Empirical divergence) is a metric designed to evaluate the distributional similarity between model-generated text and human-written reference text \citep{pillutla2021mauve}. Unlike token-level overlap metrics such as ROUGE or BLEU, MAUVE captures high-level properties of natural language, such as fluency, coherence, and diversity, by comparing the distributions of embeddings derived from pretrained language models.

To evaluate MAUVE in our setting, we use the fine-tuned model or the ICL model to generate synthetic outputs for each sample in the test set. Specifically, we input the corresponding control code from each test sample and collect the resulting generated texts. MAUVE is then computed between the set of model-generated outputs and the original texts from the test set. A higher MAUVE score indicates that the generated distribution more closely matches the distribution of human-written text, reflecting better language quality and realism.

\section{Results}
For all reported results with confidence intervals, we repeat each setting three times and compute the 95\% confidence interval based on the observed variation across runs.

We include as a baseline performing generation with ICL, where we provide the model with three examples from the training set without using any control codes. We also include DP-SGD, which clips the gradients to limit the
contribution of individual samples from the training data and subsequently adds noise from a predefined type of distribution to the sum of the clipped
gradients across all samples, as a reference for comparison with our method \cite{ramesh2025synthtexteval}.
\subsection{Private Entity Known}
The privacy metrics for all models in the private entity known setting are presented in Table~\ref{tab:private_entity_known}. In the experiment, we set epsilon = 8.

All methods achieve much higher privacy protection than the baseline. As expected, \texttt{ICL w/ privacy enhancement} achieves better (lower) PIPP and ELP than the other models. 
While our measured leakage is 0\% on average, we find that some leakage is possible in practice. LLaMA tokenizes words in a case-sensitive manner—e.g., "apple" and "Apple" are assigned different token representations. Although we construct a bad token list that accounts for multiple case variants (e.g., lowercase, uppercase), there remains a small chance of private entity leakage due to unaccounted casing variations or tokenization artifacts.
Forcing leakage to zero is still possible by repeating each generation until no private entity is detected, and since the rate of leakage is effectively 0\%, the expected number of regenerations needed is very small.  However, in our experiments, for fair comparison with other methods, we report results from a single generation pass per sample.

\begin{table*}[t!]
  \centering
  \begin{tabular}{l|cccc}
    \toprule
    \textbf{Method} & \textbf{PIPP (\%)} $\downarrow$ & \textbf{ELP (\%)} $\downarrow$ & \textbf{ROUGE-2}  $\downarrow$&\textbf{ROUGE-L} $\downarrow$\\
    \midrule
    Baseline & 50.0 $\pm$ 2.20 & 30.89 $\pm$ 3.15 & 0.3898 $\pm$ 0.02 & 0.4303 $\pm$ 0.01 \\
    ICL & 0.89 $\pm$ 0.80 & 0.74 $\pm$ 0.13 & 0.4133 $\pm$0.02& 0.4877 $\pm$ 0.01\\
    ICL w/ privacy enhancement & \textbf{0.77 $\pm$ 0.02}&  0.67$\pm$ 0.01 & 0.3968 $\pm$ 0.04 & 0.4478 $\pm$ 0.01\\
    Fine-tuning &4.24 $\pm$ 0.18 & 0.28 $\pm$ 0.03 & 0.0235 $\pm$ 0.02 & \textbf{0.0762$\pm$ 0.01}\\
    Fine-tuning w/ masking & 2.84 $\pm$ 0.02 & \textbf{0.11$\pm$ 0.03} & \textbf{0.0098 $\pm$  0.0003} & 0.083 $\pm$ 0.01\\
    \bottomrule
  \end{tabular}
  \caption{Privacy protection performance under the private entity unknown setting on the TAB dataset}
  \label{tab:private_entity_unknown}
\end{table*}

While the ICL method supports direct prevention of leakage, leading to better PIPP and ELP metrics, the higher ROUGE scores as compared to the fine-tuning methods indicate that the model copies substantially more content from the input examples. For example, ROUGE-L for ICL is 0.3073 as compared to 0.133 for fine-tuning.
This result suggests that the ICL setting is preferred when there is high confidence in the deidentification model and protecting information not specifically marked as identifiable is unimportant. In contrast, the fine-tuning approach offers a greater level of synthesis and thus is more capable of protecting information not explicitly marked as identifying but that may still be considered private. 

We also perform evaluations on the MIMIC-III dataset, which follow a similar trend to those observed on the TAB dataset and are presented in Table~\ref{mimic}.

\begin{table*}[t!]
  \centering
  \begin{tabular}{l|cccc}
    \toprule
    \textbf{Method} & \textbf{PIPP (\%)} $\downarrow$ & \textbf{ELP (\%)} $\downarrow$ & \textbf{ROUGE-2}  $\downarrow$ & \textbf{ROUGE-L} $\downarrow$ \\
    \midrule
    Baseline & 22.3 & 4.7 & 0.4557 & 0.5321 \\
    ICL & 5.6 & 1.8 & 0.3714 & 0.4147 \\
    ICL w/ privacy enhancement & \textbf{1.2} & \textbf{0.5} & 0.3921 & 0.4263 \\
    Fine-tuning & 9.7 & 2.3 & 0.1478 & 0.1526 \\
    Fine-tuning w/ masking & 4.8 & 1.2 & \textbf{0.0215} & \textbf{0.0874} \\
    \bottomrule
  \end{tabular}
  \caption{Privacy protection performance under the private entity known setting on the MIMIC-III dataset. Lower scores indicate stronger privacy protection across entity-level and lexical similarity metrics.}
  \label{mimic}
\end{table*}

\subsection{Private entity unknown}

In the entity unknown setting (\Cref{tab:private_entity_unknown}), as expected privacy is generally worse than in the private entity known setting. For the ICL setting, the PIPP and ELP scores are lower than those in the private entity known setting, primarily because in some generations, the model either failed to produce any meaningful output or generated only a few terms, leading to lower leakage rates relative to the normal outputs. The ICL w/ privacy enhancement still achieves the best PIPP, but ELP is better for the prefix fine-tuning models. This result empirically demonstrates that the greater level of synthesis in the fine-tuning models can help correct for an imperfect deidentification model. Other trends are similar, with the prefix fine-tuning models having better (lower) ROUGE scores than the ICL methods.

\subsection{Downstream Performance} To evaluate the utility of our method, we primarily report perplexity and MAUVE. As shown in Table~\ref{tab_utility}, fine-tuning achieves better utility performance compared to the ICL method, with the fine-tuning w/ masking variant performing the best overall.

In terms of privacy protection, the ICL w/ privacy enhancement achieves the strongest results on certain metrics. However, the use of bad tokens in this setting may impair utility, particularly under the private entity unknown case, where the entity matcher may incorrectly identify common terms as private. Although the baseline appears to perform well in utility metrics, it offers poor privacy protection and is therefore not a viable privacy-preserving solution.

Overall, fine-tuning with masking strikes the best balance between utility and privacy, demonstrating strong performance across both evaluation criteria.

\begin{table*}[t!]
  \centering
  \begin{tabular}{l|cc|cc}
    \toprule
    & \multicolumn{2}{c|}{\textbf{Private Entity Known}} & \multicolumn{2}{c}{\textbf{Private Entity Unknown}} \\
    \textbf{Method} & \textbf{Perplexity } $\downarrow$ & \textbf{MAUVE} $\uparrow$ & \textbf{Perplexity} $\downarrow$ & \textbf{MAUVE} $\uparrow$ \\
    \midrule
    Baseline & 11.7 & 0.78 & \textbf{11.7} & \textbf{0.78} \\
    DP-SGD & 12.8 & 0.70 & 12.8 & 0.70\\
    ICL & 11.8 & 0.78 & 13.5 & 0.71 \\
    ICL w/ privacy enhancement & 12.0 & 0.66 & 14.2 & 0.62 \\
    Fine-tuning & 10.5 & 0.82 & 12.1 & 0.76 \\
    Fine-tuning w/ masking & \textbf{10.2} & \textbf{0.83} & \textbf{11.7} & \textbf{0.78} \\
    \bottomrule
  \end{tabular}
  \caption{Utility performance under the private entity known and unknown settings on the TAB dataset}
  \label{tab_utility}
\end{table*}

\subsection{Ablation Study}
\subsubsection{Performance on larger model}
All experiments described above were conducted using Sheared-LLaMA 1.3B. We extended our evaluation to Llama 3.1-8B Instruct, testing both ICL and ICL with privacy enhancement, shown in \Cref{tab:large_model}. The overall trends remain consistent: both settings show low risk of leakage, with the privacy enhancement setting providing greater protection. This demonstrates our method's robust performance across different model scales.

While PIPP and ELP metrics indicate slightly higher leakage risk in the larger model, we attribute this to the fact that TAB is not truly a private dataset. Larger LLMs exposed to TAB during pre-training are more likely to have memorized its content. Notably, Rouge-L scores improved compared to the smaller model, suggesting greater degrees of synthesizing at larger scales.

\begin{table*}[t!]
  \centering
  \begin{tabular}{l|cccc}
    \toprule
    \textbf{Method} & \textbf{PIPP (\%)} $\downarrow$ & \textbf{ELP (\%)} $\downarrow$ & \textbf{ROUGE-2}  $\downarrow$ & \textbf{ROUGE-L} $\downarrow$ \\
    \midrule
    ICL & 6.51 & 2.50 & 0.2663 & 0.3201 \\
    ICL w/ privacy enhancement & 3.55 & 0.82 & 0.2793 & 0.3341 \\
    \bottomrule
  \end{tabular}
  \caption{Privacy protection performance under the private entity known setting on the TAB dataset for a larger LLM (Llama 3.1-8B Instruct). Lower scores indicate stronger privacy protection across entity-level and lexical similarity metrics.}
  \label{tab:large_model}
\end{table*}
\subsubsection{Performance on both direct and quasi identifiers}

In the previous experiments, we considered only direct identifiers as private entities. We further evaluate the scenario where both direct and quasi-identifiers are treated as private. As shown in Table~\ref{direct_quasi}, performance generally declines due to the significant increase in the number of privacy terms. Nonetheless, the ICL w/ privacy enhancement setting continues to achieve the best results in the PIPP and ELP metrics, demonstrating the effectiveness of the bad token strategy in preventing privacy leakage.

For ROUGE-2 and ROUGE-L, the prefix tuning with masking variant remains the strongest performer. Compared to the setting where only direct identifiers are protected, its privacy protection performance shows only a modest decline, especially relative to other methods. This result highlights the method’s robustness and its potential to scale effectively to scenarios involving a larger set of sensitive entities. The performance on pure quasi identifiers is shown in Appendix~\ref{quasi_identifiers}.

\begin{table*}[t!]
  \centering
  \begin{tabular}{l|cccc}
    \toprule
    \textbf{Method} & \textbf{PIPP (\%)} $\downarrow$ & \textbf{ELP (\%)} $\downarrow$ & \textbf{ROUGE-2}$\downarrow$ &\textbf{ROUGE-L} $\downarrow$ \\
    \midrule
    Baseline & 66.70 $\pm$ 5.22 & 48.2 $\pm$ 4.17 & 0.3898 $\pm$ 0.02 & 0.4303 $\pm$ 0.01 \\
    ICL & 27.31 $\pm$ 6.18 & 3.20 $\pm$ 1.13 & 0.2669 $\pm$ 0.009 &  0.3034 $\pm$ 0.01 \\
    ICL w/ privacy enhancement  & \textbf{0.89 $\pm$ 0.332}  & \textbf{0.06 $\pm$ 0.027} & 0.2624 $\pm$ 0.03 & 0.2988 $\pm$ 0.004\\
    Fine-tuning & 2.74 $\pm$ 1.91 & 0.64 $\pm$ 0.16 & 0.012 $\pm$ 0.0018 & 0.11 $\pm$ 0.08\\
    Fine-tuning w/ masking & 3.17 $\pm$ 0.99 & 0.33 $\pm$ 0.09& \textbf{0.009 $\pm$ 0.001} & \textbf{0.0977 $\pm$ 0.01}\\
    \bottomrule
  \end{tabular}
  \caption{Privacy protection performance under the private entity known setting, where both direct and quasi-identifiers are treated as private entities, on the TAB dataset }
  \label{direct_quasi}
\end{table*}

%% file: paper/appendix.tex
\appendix

\section{Control Code}
\subsection{Control Code of TAB Dataset}
\label{appendix_tab_control_code}
The control codes in the TAB dataset are categorized into eight types, as shown in Table~\ref{tab:private_entity_categories}.

\begin{table*}[t!]
\centering
\begin{tabular}{|l|p{13cm}|}
\hline
\textbf{Category} & \textbf{Description} \\
\hline
\textbf{PERSON} & Names of individuals, including full names, nicknames, aliases, usernames, and initials. \\
\hline
\textbf{CODE} & Identifying numbers and codes such as Social Security Numbers (SSNs), phone numbers, passport numbers, license plates, and other personal identifiers. \\
\hline
\textbf{LOC} & Locations and addresses, including cities, regions, countries, streets, and named infrastructures (e.g., bus stops, bridges). \\
\hline
\textbf{ORG} & Names of organizations such as companies, educational institutions, government bodies, healthcare facilities, non-governmental organizations, and religious institutions. \\
\hline
\textbf{DEM} & Demographic attributes including ethnicity, language, heritage, job titles, ranks, education levels, physical descriptions, medical diagnoses, birthmarks, and age-related information. \\
\hline
\textbf{DATETIME} & Mentions of specific dates (e.g., ``October 3, 2018''), times (e.g., ``9:48 AM''), or durations (e.g., ``18 years''). \\
\hline
\textbf{QUANTITY} & Quantitative values such as percentages, measurements, or monetary amounts. \\
\hline
\textbf{MISC} & Any other information that could describe an individual but does not fall under the above categories. \\
\hline
\end{tabular}
\caption{Private entity categories and their descriptions used for privacy annotation.}
\label{tab:private_entity_categories}
\end{table*}

\subsection{Fictional Control Code}
\label{fictional_control_code}
During generation, we create fictional control codes that follow the format and structure of the original context. The types of fictional control codes are determined based on the categories present in the input context. Specifically, we randomly generate values for seven predefined categories—excluding \texttt{MISC}—to simulate plausible yet non-identifiable entities. The generation rules and sampling patterns for each category are detailed in Figure~\ref{fig:fictional_control_codes}.
\begin{figure*}[t!]
\centering
\small
\begin{tabular}{|l|p{12cm}|}
\hline
\textbf{Control Code Category} & \textbf{Random Generation Pattern and Sample Values} \\
\hline
\texttt{CODE} & Random alphanumeric format: \texttt{ABCDE/XY} \newline Example: \texttt{X5T9L/QZ} \\
\hline
\texttt{PERSON} & Title + First Name + Last Name \newline Sample pool: Titles = \texttt{Mr, Ms, Dr, Prof}, First Names = \texttt{Alex, Blake, Casey, Dana, Elliot, 
Finley, Harper, Jordan, Kai, Logan, Morgan, Quinn, Riley, Skyler}, Last Names = \texttt{Adams, Baker, Carson, Dawson, Ellis, Foster, Griffin, Hayes, Irwin, Johnson, Kennedy, Lewis} \newline Example: \texttt{Dr Logan Ellis} \\
\hline
\texttt{DATETIME} & Random date between 1990–2024 in the format: \texttt{DD Month YYYY} \newline Example: \texttt{12 October 2011} \\
\hline
\texttt{LOC} & Random choice from: \newline - Cities: \texttt{Baltimore, Seattle, Tokyo, Munich, Cairo} \newline - Countries: \texttt{USA, Germany, Japan, Kenya, Brazil} \newline - Addresses: \texttt{221B Baker St, 1600 Amphitheatre Pkwy, 350 Fifth Ave} \newline - Infrastructure: \texttt{London Bridge, Central Station, Pier 39} \newline Example: \texttt{1600 Amphitheatre Pkwy} \\
\hline
\texttt{ORG} & Sampled from organization names: \texttt{OpenAI, World Health Organization, Harvard University, UNICEF, St. Mary’s Hospital, SpaceX, NASA, MIT, Stanford University, Google} \newline Example: \texttt{Stanford University} \\
\hline
\texttt{DEM} & Pattern-based combinations of: \newline - Heritage: \texttt{Irish-American, Nigerian, Chinese, Latinx, Punjabi} \newline - Jobs: \texttt{software engineer, nurse, professor, mechanic, pilot} \newline - Ages: randomly generated as \texttt{N-year-old} \newline Example: \texttt{42-year-old pilot} or \texttt{Latinx descent} \\
\hline
\texttt{QUANTITY} & Randomly selected format: \newline - Percentage: \texttt{45\%} \newline - Currency: \texttt{\$87,500} \newline Example: \texttt{\$215,000} \\
\hline
\end{tabular}
\caption{Pattern-based generation of fictional control code values for each privacy entity category. Each value is randomly sampled using predefined lists and templates to ensure diversity while preserving format consistency.}
\label{fig:fictional_control_codes}
\end{figure*}
\section{Rouge Score}
\label{rouge_score}
ROUGE-2 measures the overlap of bigrams between the generated and reference texts, capturing local phrase-level similarity. Let $|B_{\text{match}}|$ be the number of overlapping bigrams. Then, the precision $P$ and recall $R$ are given by:
\[
P = \frac{|B_{\text{match}}|}{|B_{\text{gen}}|}, \quad R = \frac{|B_{\text{match}}|}{|B_{\text{ref}}|}
\]
The ROUGE-2 F1 score is computed as the harmonic mean:
\[
\text{ROUGE-2} = \frac{2 \cdot P \cdot R}{P + R}
\]

ROUGE-L, on the other hand, computes the length of the longest common subsequence (LCS), which reflects the preservation of word order and sentence structure. 
Let $\text{LCS}(X, Y)$ denote the length of the longest common subsequence between a reference $X$ and a generation $Y$, and let $|X|$ and $|Y|$ be the lengths of the reference and generation, respectively. Then:
\[
P = \frac{\text{LCS}(X, Y)}{|Y|}, \quad R = \frac{\text{LCS}(X, Y)}{|X|}
\]
The ROUGE-L F1 score is given by:
\[
\text{ROUGE-L} = \frac{(1 + \beta^2) \cdot P \cdot R}{R + \beta^2 \cdot P}, \quad \text{where } \beta = 1
\]

\section{Baseline}
The format of baseline ICL in shown in Table~\ref{baseline}.

\label{in-context-learning}
\begin{table*}[t!]
\centering
\renewcommand{\arraystretch}{1.4}
\begin{tabular}{|p{14cm}|}
\hline
\multicolumn{1}{|c|}{\textbf{In-Context Learning Prompt to LLM}} \\
\hline
\multicolumn{1}{|c|}{\textbf{First Context}} \\
\hline

The case originated in an application (no. 36244/06) against the Kingdom of Denmark lodged with the Court under Article 34 of the Convention for the Protection of Human Rights and Fundamental Freedoms (“the Convention”) by a Danish national, Mr Henrik Hasslund (“the applicant”), on 31 August 2006.
The applicant was represented by Mr Tyge Trier, a lawyer practising in Copenhagen. The Danish Government (“the Government”) were represented by their Agent, Ms Nina Holst-Christensen of the Ministry of Justice.
\\
\hline
\multicolumn{1}{|c|}{\textbf{Second Context}} \\
\hline

The case originated in an application (no. 29366/03) against the Republic of Poland lodged with the Court under Article 34 of the Convention for the Protection of Human Rights and Fundamental Freedoms (“the Convention”) by Mr D. Stępniak on 25 July 2003.
The Polish Government (“the Government”) were represented by their Agent, Mr J. Wołąsiewicz of the Ministry of Foreign Affairs. \\
\hline
\multicolumn{1}{|c|}{\textbf{Third Context}} \\
\hline

The case originated in an application (no. 5138/04) against the Republic of Turkey lodged with the Court under Article 34 of the Convention for the Protection of Human Rights and Fundamental Freedoms (“the Convention”) by a Turkish national, Mr Nusret Amutgan (“the applicant”), on 29 December 2003.
The applicant was represented by Ms B Özpolat, a lawyer practising in Gaziantep. The Turkish Government (“the Government”) were represented by their Agent.
 \\
\hline

\end{tabular}
\caption{Example of a 3-shot baseline in-context learning prompt used to generate a single synthetic passage.}
\label{baseline}
\end{table*}

\subsection{Generation}
An example of a generated output using the baseline ICL method based on the above prompt is shown in Table~\ref{baseline_example}.

\begin{table*}[t!]
\centering
\small
\begin{tabular}{|p{15cm}|}
\hline
\textbf{Example generation from Baseline} \\
\hline

\texttt{PROCEDURE} \\
The case originated in an application (no. 5138/04) against the Republic of Turkey lodged with the Court under Article 34 of the Convention for the Protection of Human Rights and Fundamental Freedoms (“the Convention”) by Mr Nusret Amutgan (“the applicant”), on 29 December 2003. \\
The applicant was represented by Ms B Özpolat, a lawyer practising in Gaziantep. The Turkish Government (“the Government”) were represented by their Agent. \\

\vspace{1mm}

\texttt{PROCEDURE} \\
The case originated in an application (no. 5138/04) against the Republic of Turkey lodged with the Court under Article 34 of the Convention for the Protection of Human Rights and Fundamental Freedoms (“the Convention”) by Mr Nusret Amutgan (“the applicant”), on 29 December 2003. \\
The applicant was represented by Ms B Özpolat, a lawyer practising in Gaziantep. The Turkish Government (“the Government”) were represented by their Agent. \\

\hline
\end{tabular}
\caption{Example generation of baseline in-context learning }
\label{baseline_example}
\end{table*}

\section{In-context learning}
The format of input of in-context learning is shown in Table~\ref{tab:vertical_prompt_structure}.
\begin{table*}[t!]
\centering
\renewcommand{\arraystretch}{1.4}
\begin{tabular}{|p{15cm}|}
\hline
\multicolumn{1}{|c|}{\textbf{In-Context Learning Prompt to LLM}} \\
\hline
\multicolumn{1}{|c|}{\textbf{First Context}} \\
\hline
CODE: 36244/06\\
PERSON: Mr Henrik Hasslund, Mr Tyge Trier, Ms Nina Holst-Christensen\\
DATETIME: 31 August 2006\\
PROCEDURE\\
The case originated in an application (no. 36244/06) against the Kingdom of Denmark lodged with the Court under Article 34 of the Convention for the Protection of Human Rights and Fundamental Freedoms (“the Convention”) by a Danish national, Mr Henrik Hasslund (“the applicant”), on 31 August 2006.
The applicant was represented by Mr Tyge Trier, a lawyer practising in Copenhagen. The Danish Government (“the Government”) were represented by their Agent, Ms Nina Holst-Christensen of the Ministry of Justice.
\\
\hline
\multicolumn{1}{|c|}{\textbf{Second Context}} \\
\hline
CODE: 29366/03\\
PERSON: Mr D. Stępniak, Mr J. Wołąsiewicz\\
DATETIME: 25 July 2003\\
PROCEDURE\\
The case originated in an application (no. 29366/03) against the Republic of Poland lodged with the Court under Article 34 of the Convention for the Protection of Human Rights and Fundamental Freedoms (“the Convention”) by Mr D. Stępniak on 25 July 2003.
The Polish Government (“the Government”) were represented by their Agent, Mr J. Wołąsiewicz of the Ministry of Foreign Affairs. \\
\hline
\multicolumn{1}{|c|}{\textbf{Third Context}} \\
\hline
CODE: 5138/04\\
PERSON: Mr Nusret Amutgan, Ms B Özpolat\\
DATETIME: 29 December 2003\\
PROCEDURE\\
The case originated in an application (no. 5138/04) against the Republic of Turkey lodged with the Court under Article 34 of the Convention for the Protection of Human Rights and Fundamental Freedoms (“the Convention”) by a Turkish national, Mr Nusret Amutgan (“the applicant”), on 29 December 2003.
The applicant was represented by Ms B Özpolat, a lawyer practising in Gaziantep. The Turkish Government (“the Government”) were represented by their Agent.
 \\
\hline
\multicolumn{1}{|c|}{\textbf{Randomly Generated Control Code}} \\
\hline
CODE: JMIRM/QT\\
PERSON: Ms Elliot Carson\\
DATETIME: 16 June 2007\\
\hline
\end{tabular}
\caption{Example of a 3-shot in-context learning prompt used to generate a single synthetic passage.}
\label{tab:vertical_prompt_structure}
\end{table*}

\subsection{Generation}
\begin{table*}[t!]
\centering
\small
\begin{tabular}{|p{13.5cm}|}
\hline
\textbf{Example generation from ICL} \\
\hline
\textbackslash n\texttt{PROCEDURE}\textbackslash n The case originated in an application (no. 29762/99) against the Republic of Yugoslavia lodged with the Court under Article 34 of the Convention for the Protection of Human Rights and Fundamental Freedoms (“the Convention”) by a Yugoslav national, Prof Kai Kennedy (“the applicant”), on 20 January 2000. The applicant was represented by Mr J. B. S. Fleming, a lawyer practising in Zagreb. The Republic of Yugoslavia (“the Republic”) were represented by their Agent, Ms I. K. Jankovic of the Ministry of Foreign Affairs. \\
\hline
\end{tabular}
\caption{Example paragraph generated from the above prompt using the in-context learning (ICL) method}
\label{icl_example}
\end{table*}

An example of a generated output using the in-context learning (ICL) method based on the above prompt is shown in Table~\ref{icl_example}.

\section{ICL w/ privacy enhancement}
The input format follows the same structure used for in-context learning, as illustrated in Table~\ref{tab:vertical_prompt_structure}.

\subsection{Generation}

An example of a generated output using the ICL w/ privacy enhancement method based on the above prompt is shown in Table~\ref{icl_privacy_example}.

\begin{table*}[t!]
\centering
\small
\begin{tabular}{|p{13.5cm}|}
\hline
\textbf{Generated Paragraph from ICL w/ privacy enhancement} \\
\hline
\texttt{PROCEDURE} \\
The case originated in an application (no. 2AGVC/7B) against the Republic of Turkey lodged with the Court under Article 34 of the Convention for the Protection of Human Rights and Fundamental Freedoms (“the Convention”) by Professor Skyler Baker (“the applicant”), on 26 August 2006. \\
The applicant was represented by Mr M. S. Yildirim, a lawyer practising in Istanbul. The Turkish Government (“the Government”) were represented by their Agent, Ms Z. Özkan of the Ministry of Foreign Affairs. \\
\hline
\end{tabular}
\caption{Example paragraph generated using icl w/ privacy enhancement method.}
\label{icl_privacy_example}
\end{table*}

\section{Prefix-tuning}
The prompt of prefix-tuning will directly be the fictional code. Generally, the final paragraph will be cut off for the downstream performance. 
\subsection{Generation}
An example of a generated output using the prefix tuning method is shown in Table~\ref{prefix_example}.
\begin{table*}[t!]
\centering
\small
\begin{tabular}{|p{13.5cm}|}
\hline
\textbf{Example Paragraph of prefix-tuning} \\
\hline
LOCATION: DUYNK\\
Skyler Johnson was 19 years old when he died in a car crash on the night of 9 September 2018.\\
He was in the passenger seat of a grey Holden Commodore which collided with a white Toyota Corolla on the night of 9 September 2018.\\
The Corolla driver, 38 year old, was not injured.\\
\hline
\end{tabular}
\caption{Full academic biography paragraph and publication entries used as example input text.}
\label{prefix_example}
\end{table*}

\section{Performance on quasi identifiers}
\label{quasi_identifiers}
Beyond experiments on direct identifiers and combined direct and quasi-identifier settings, we evaluated performance on quasi-identifiers alone.

Figure~\ref{pure_quasi1} shows results for quasi-identifiers under the private known setting. As expected, privacy-related tasks show some performance degradation due to their inherent difficulty. The ICL method achieves approximately 31\% on PIPP metrics. In contrast, both finetuning and finetuning with masking demonstrate only modest performance drops, confirming their effectiveness for privacy protection. Notably, ICL with privacy enhancement provides strong protection, achieving near-zero scores on both PIPP and ELP metrics. However, this method maintains higher Rouge-2 and Rouge-L scores compared to finetuning approaches, indicating excessive similarity to original articles that could pose privacy risks.

Figure~\ref{pure_quasi2} presents results under the private unknown setting. ICL with privacy enhancement shows increased leakage (10\% in PIPP, 3\% in ELP) because Presidio's entity recognition algorithm struggles to identify quasi-identifiers. Consequently, many of the identifiers are not set to zero likelihood, resulting in increased privacy leakage. The masking approach is ineffective due to poor quasi-identifier detection. However, the finetuning stage introduces sufficient generation diversity to limit privacy leakage to approximately 7\%.
\begin{table*}[t!]
  \centering
  \begin{tabular}{l|cccc}
    \toprule
    \textbf{Method} & \textbf{PIPP (\%)} $\downarrow$ & \textbf{ELP (\%)} $\downarrow$ & \textbf{ROUGE-2}$\downarrow$ &\textbf{ROUGE-L} $\downarrow$ \\
    \midrule
    ICL & 31.36 & 3.94 & 0.2726 & 0.3057 \\
    ICL w/ privacy enhancement  & \textbf{0.88} & \textbf{0.06} & 0.2624 & 0.2988 \\
    Fine-tuning & 2.76 & 1.02  & 0.0087  & 0.0794\\
    Fine-tuning w/ masking & 2.88 & 1.07 & \textbf{0.0084} & \textbf{0.0793}\\
    \bottomrule
  \end{tabular}
  \caption{Privacy protection performance under the private entity known setting, where only quasi identifiers are treated as private entities, on the TAB dataset }
  \label{pure_quasi1}
\end{table*}

\begin{table*}[t!]
  \centering
  \begin{tabular}{l|cccc}
    \toprule
    \textbf{Method} & \textbf{PIPP (\%)} $\downarrow$ & \textbf{ELP (\%)} $\downarrow$ & \textbf{ROUGE-2}$\downarrow$ &\textbf{ROUGE-L} $\downarrow$ \\
    \midrule
    ICL & 40.21 & 6.79 & 0.3233 & 0.0.3487 \\
    ICL w/ privacy enhancement  & 13.21 & 3.06 & 0.2624 & 0.2988 \\
    Fine-tuning & \textbf{7.21} & 3.02  & 0.053  & 0.0981\\
    Fine-tuning w/ masking & 7.34 & \textbf{2.38} & \textbf{0.019} & \textbf{0.0823}\\
    \bottomrule
  \end{tabular}
  \caption{Privacy protection performance under the private entity unknown setting, where only quasi identifiers are treated as private entities, on the TAB dataset }
  \label{pure_quasi2}
\end{table*}